\documentclass{article}

\usepackage{microtype}
\usepackage{graphicx}
\usepackage{subfigure}
\usepackage{booktabs} 
\usepackage{multirow}
\usepackage{graphicx}      
\usepackage{subcaption}    

\usepackage{hyperref}

\usepackage[accepted]{icml2024}

\usepackage{amsmath}
\usepackage{amssymb}
\usepackage{mathtools}
\usepackage{amsthm}
\usepackage[capitalize,noabbrev]{cleveref}

\theoremstyle{plain}

\theoremstyle{definition}

\theoremstyle{remark}

\usepackage[textsize=tiny]{todonotes}

\icmltitlerunning{From Graph Diffusion to Graph Classification}

\newcommand{\ie}[1]{\textit{i.e.} #1}

\newcommand{\clfLoss}{L_\text{CLF}}
\newcommand{\sumLoss}{L_\text{SUM}}
\newcommand{\denLoss}{L_\text{DEN}}

\DeclareMathOperator{\sft}{Softmax}
\begin{document}

\twocolumn[
\icmltitle{From Graph Diffusion to Graph Classification}




\begin{icmlauthorlist}
\icmlauthor{Jia Jun Cheng Xian}{ubc}
\icmlauthor{Sadegh Mahdavi}{ubc}
\icmlauthor{Renjie Liao}{ubc}
\icmlauthor{Oliver Schulte}{sfu}

\end{icmlauthorlist}

\icmlaffiliation{ubc}{Department of Electrical and Computer Engineering, University of British Columbia, Vancouver, Canada}
\icmlaffiliation{sfu}{Department of Computer Science, Simon Fraser University, Burnaby, Canada}

\icmlcorrespondingauthor{Jia Jun Cheng Xian}{anthony@ece.ubc.ca}

\icmlkeywords{Machine Learning, ICML}

\vskip 0.3in
]



\printAffiliationsAndNotice{\icmlEqualContribution} 

\begin{abstract}

Generative models such as diffusion models have achieved remarkable success in state-of-the-art image and text tasks. Recently, score-based diffusion models have extended their success beyond image generation, showing competitive performance with discriminative methods in image {\em classification} tasks~\cite{zimmermann2021score}. However, their application to classification in the {\em graph} domain, which presents unique challenges such as complex topologies, remains underexplored. We show how graph diffusion models can be applied for graph classification. We find that to achieve competitive classification accuracy, score-based graph diffusion models should be trained with a novel training objective that is tailored to graph classification. In experiments with a sampling-based inference method, our discriminative training objective achieves state-of-the-art graph classification accuracy.
\end{abstract}

\section{Introduction}
\label{submission}

Recent breakthroughs with generative models have enabled outstanding performance in challenging tasks in various modalities such as image generation~\cite{imagen}, speech generation~\cite{le2023voicebox}, and natural language processing~\cite{openai2024gpt4}. For text, early GPT models~\cite{gpt1, gpt2} showed that generative training not only provides excellent text generation performance, but can be competitive to models discriminatively trained in the downstream task. Later improvements showed that generative training significantly outperforms all alternative approaches~\cite{gpt3, openai2024gpt4}. In the image domain, \citet{secret_zero_shot} show that text-to-image diffusion models hold promise for {\em zero-shot classification on images} without any additional discriminative training. \citet{zimmermann2021score} train score-based diffusion models for zero-shot classification and show competitive performance with SOTA discriminative models in the CIFAR-10 data set. Collectively, these previous works show how sophisticated generative models can be leveraged for 
classification tasks.


While there has been extensive research on the applicability of generative models for the classification tasks on image and text domains, this question has remained unexplored in the {\em graph domain.} Therefore, it is natural to ask: \emph{Do generative models bring the same classification competitiveness to graphs?} This work takes a step toward answering this question: we show that generative models are indeed strong baselines for graph classification. In particular, our contributions are as follows.

\begin{itemize}
    \item We find that purely generatively trained graph diffusion models do not perform well as zero-shot classifiers using exact likelihood for inference. We therefore develop a new discriminative training objective, based on the generative ELBO likelihood approximation, that leads to strong classification performance as well as high-quality graph generation.
    \item Our base diffusion model is not permutation invariant. We show that both training and inference can be improved by randomly sampling adjacency matrices from the isomorphism class of the training/test graph.
    \item 
    We conduct extensive experiments on three training loss objectives, each paired with two different inference methods, using the K-Regular synthetic dataset and the real-world IMDB-BINARY and PROTEIN datasets. We identify an optimal combination of training objectives and inference methods that is efficient and achieves superior performance across our baselines.
\end{itemize}

\section{Related Work}
\textbf{Graph Neural Networks and Graph Classification.}
Graph Neural Networks (GNNs) (\citep{dgcnn, gin,graphSAGE,diffpool,ecc}) have emerged as effective architectures to process graph-structured data. The core operation of a GNN layer involves updating the representation of each node. The new node representation is updated by aggregating the information from the neighbor of each node and its own node representation. The final node representations can be used for downstream tasks such as graph classification. \citet{Errica2019AFC} provided a fair comparison of various graph classifiers (\cite{dgcnn, gin,graphSAGE,diffpool,ecc}). We adopt their evaluation process and framework (details in \Cref{sec: experiment} and Appendix \ref{App: experiment}).

\textbf{Diffusion Models.}
Diffusion models (\citet{song2021scorebased, ddpm, vdm, edm, rombach2021highresolution}) are generative model that operate through a dual process comprising forward and backward steps. The forward process involves introducing noise into the data, with the noise level denoted by $\sigma(t)$ at each time step $t$. Conversely, the backward process aims to denoise the data, transitioning from a noisy state to a clean one. Variational Diffusion Models \cite{ddpm, vdm, edm, rombach2021highresolution} model the discrete time backward process by learning the evidence lower bound (ELBO) of the backward process. \citet{song2021scorebased} utilize continuous-time diffusion, based on Stochastic Differential Equations (SDEs), and demonstrate that the data likelihood function of a scored-based diffusion model can be accurately estimated by solving an ordinary differential equation. This type of diffusion has demonstrated significant success, particularly in the field of image processing. 

\textbf{Graph generative diffusion models.}
Motivated by the success of diffusion models on image generation, several studies have been conducted to extend these models to the graph domain (\citet{gdss, vignac2023digress, auto_graph_diffusion, grand}). \citet{vignac2023digress} propose an equivariant discrete denoising diffusion model to generate graphs. \citet{auto_graph_diffusion} use a node-absorbing diffusion process for discrete graph space generation. Other graph diffusion models (\citep{diffdock, difflinked, geodiff}) focus on molecular data and achieve remarkable results. \citet{yan2023swingnn} 
introduce SwinGNN, a non-equivariant score-based diffusion model that achieves state-of-the-art results on several benchmarks. Given the strong generation performance of SwinGNN, we use this architecture as our backbone graph generative model, and show how to adapt the SwinGNN model for classification tasks. 


\textbf{Generative Classifiers.} For non-graph i.i.d. data, it is well known that a classification model can be derived from a generative model~\cite{Bishop2006}. The advantages and disadvantages of a generative approach to classification have been extensively researched, both theoretically and empirically~\cite{Ng2002}. A key finding is that generative classifiers approach their maximum accuracy faster with fewer datapoints than discriminative classifiers. 
%
Because graph data are much more complex than i.i.d. data, much work has gone into developing sophisticated generative models for graphs. It is desirable to leverage these models for graph classification. 
To our knowledge, ours is the first work to study the generative approach for graph classification. 


\textbf{Diffusion generative classifier.}
One line of research studied how to adapt diffusion models for classification tasks in the {\em image domain.} \citet{zimmermann2021score} add the class label as a conditioning variable to score-based diffusion models and take advantage of the fact that the exact computation of likelihood is possible for score-based models to perform image classification tasks. \citet{secret_zero_shot} leverage trained text-to-image diffusion models (such as Stable Diffusion~\cite{rombach2021highresolution}) to perform zero-shot classification. This method relies on estimating the class-conditional likelihood 
by computing an evidence lower bound (ELBO) on an image and its candidate labels. Our work is related to both works since we consider diffusion models for classification. It is different in 1) the modality we study (graph vs. image/text), 2) the loss function we use to train the diffusion model.

\section{Background on Graph and Diffusion Models} 

We introduce background and notation for graphs and for graph diffusion models.


\subsection{Graph}

A graph is a collection of nodes (or vertices) and edges connecting pairs of nodes, often used to model relationships and structures in various scientific fields or social networks. In this work, we assume that all graphs are undirected and unweighted. Formally, a graph \( G \) is defined as an ordered pair \( G = (V, E) \):
\( V \) is the set of vertices, such as \( V = \{v_1, v_2, \ldots, v_n\} \).
\( E \) is the set of edges, where each edge is an unordered pair \( \{v_i, v_j\} \) in the case of an undirected graph.

The adjacency matrix \( A \) of a graph \( G \) is an \( n \times n \) matrix where \( n \) is the number of nodes. The element \( A_{ij} \) is 1 if there is an edge between the nodes \( v_i \) and \( v_j \), and 0 otherwise.
The node attribute matrix \( X \) is an \( n \times d \) matrix where each row corresponds to a node and the columns represent \( d \) attributes associated with that node.
A permutation of a graph involves rearranging the nodes of the graph according to some permutation. This is represented mathematically by a permutation matrix \( P \), a square \( n \times n \) matrix where each row and each column contains exactly one entry of 1 and all other entries are 0. The permuted adjacency matrix \( A' \) and the permuted node attribute matrix \( X' \) is given by: $A' = PAP^T$ and  $X' = PX$. 
These transformations ensure that the structural and attribute properties of the graph are preserved under the node permutation, providing a powerful tool for analyzing graph isomorphisms and symmetries.

\subsection{Diffusion Models} \label{sec: diffusion_model}
Diffusion model involves a forward process  and a reverse process. The forward process adds time-dependent noise to data, which transforms the ground true unknown distribution $q(\bold{x}_0)$, through a Markov process ${\bold{x}_0, \bold{x}_1, \ldots, \bold{x}_{T-1}}$, into a easy-to-sample distribution $q(\bold{x}_T)$, usually multivariate standard Gaussian. The ground truth distribution is thus modelled as 
\begin{equation}
\label{eq:true_q}
    q(\bold{x}_0) = \int{q(\bold{x}_T) \Pi^{T}_{t=1} q(\bold{x}_{t-1}|\bold{x}_t) d\bold{x}_{1:T}}. 
\end{equation}
There are two basic types of diffusion models, which different motivations, but essentially equivalent final forms. A Variational Diffusion Model (VDM) models $q(\bold{x}_0)$ with a forward process $q(\bold{x}_{t-1}|\bold{x}_t)$ and a learned reverse process $p_\theta(\bold{x}_{t-1}|\bold{x}_t)$. The evidence lower bound (ELBO) associated with a VDM
\begin{equation}
    \log q(\bold{x}_0) \geq \mathbb{E}_{q} \left[ \log \frac{p_\theta(\bold{x}_{0:T})}{q(\bold{x}_{1:T} | \bold{x}_0)} \right].
\end{equation}
Following previous work, this ELBO can be transformed into the {\em training loss objective} 
\begin{equation}
\label{eq:denoise_loss_x}
    \mathbb{E}_{q(\bold{x}_0),t \sim U[0,1],p_{\sigma(t)}(\Tilde{\bold{x}}|\bold{x})} \left[ \left\Vert D_{\theta}(\Tilde{\bold{x}}, t) - \bold{x} \right\Vert_{F}^{2} \right],
\end{equation}
where $q(\bold{x}_0)$ is the ground truth data distribution, $\Tilde{\bold{x}}$ is the noisy data, $p_{\sigma(t)}(\Tilde{\bold{x}}|\bold{x})$ is the noisy data distribution at time step $t$ under noise level function $\sigma(t)$, the notation $D_{\theta}(\Tilde{\bold{x}}, t)$ denotes the denoising  network and $||.||_F$ is the Frobenius norm. 

The other type are \textbf{Score-based} models \cite{song2021scorebased}, which ususally define the forward and backward process via stochastic differential equations (SDEs).
The equations usually have the following form:
\begin{equation} \label{eq: x_forward}
    d\bold{x}_t^{+} = f(\bold{x}_t, t) \, dt + g(\bold{x}_t, t) \, dW_t
\end{equation}
\begin{equation} \label{eq: x_backward}
    d\bold{x}_t^{-} = [-f(\bold{x}_t, t) + g(\bold{x}_t, t)^2 \nabla_{\bold{x}_t} \log p_t(\bold{x}_t)] \, dt + g(\bold{x}_t, t) \, dW_t
\end{equation}
where $d\bold{x}_t^{+}$ refers to the forward process, $d\bold{x}_t^{-}$ refers to the backward process, \( f(\bold{x}_t, t) \) is the drift coefficient, \( g(\bold{x}_t, t) \) is the diffusion coefficient, and \( dW_t \) denotes the standard Wiener process. It learns to approximate the gradient of the log-probability of the denoising process. The {\em training loss objective} for score-based diffusion models is defined as follows:
\begin{equation} \label{eq:scored_based_loss}
     \mathbb{E}_{q(\bold{x}_0), q(\bold{x}_t | \bold{x}_0)} \left[ \left\Vert \nabla_{\bold{x}_t} \log q(\bold{x}_t|\bold{x}_0) - s_\theta(\bold{x}_t, t) \right\Vert^2 \right]
\end{equation}
where $s_\theta(\bold{x}_t, t)$ is the score function estimated by the model at time $t$, and $q(\bold{x}_t | \bold{x}_0)$ represents the distribution of the noisy data at time $t$ conditioned on the original data $\bold{x}_0$. This is an instance of a score-based objective that aims to estimate the gradient of the log probability density of the data at various levels of noise, facilitating the generation of samples by guiding the reverse diffusion process. Theoretically,  optimizing the denoising loss objective function almost surely leads to an optimal score network \citep{6795935}. \\
Moreover, \citet{song2021scorebased} demonstrate that every SDEs has a deterministic probability flow ODE such that its trajectories share the same marginal probability density as the SDE. In other words, a diffusion model can accurately estimate its likelihood by solving an ordinary differential equation. In practice, people use an ODE solver to approximate the solution. Mathematically, the data likelihood function is given by \cite{song2021scorebased, zimmermann2021score}
\begin{equation}
\label{eq:conditioned_loglikelihood}
\begin{aligned} 
   \log p(\bold{x}_0) &= \log p_T \left( \bold{x}_0 + \int_0^T \tilde{f}_\theta(\bold{x}_t, t) \, dt \right) + \int_0^T \nabla \cdot \tilde{f}_\theta(\bold{x}_t, t) \, dt,
\end{aligned}
\end{equation}
with 
\begin{equation}
\label{eq:conditioned_f}
   \tilde{f}_\theta(\bold{x}_t, t) = \frac{\bold{x}_t - D_{\theta}(\bold{x}_t, t)}{t},
\end{equation}


\subsection{Graph diffusion Model} \label{sec: swinGNN}

\begin{figure*}
    \centering
    \includegraphics[width=1.0\textwidth]{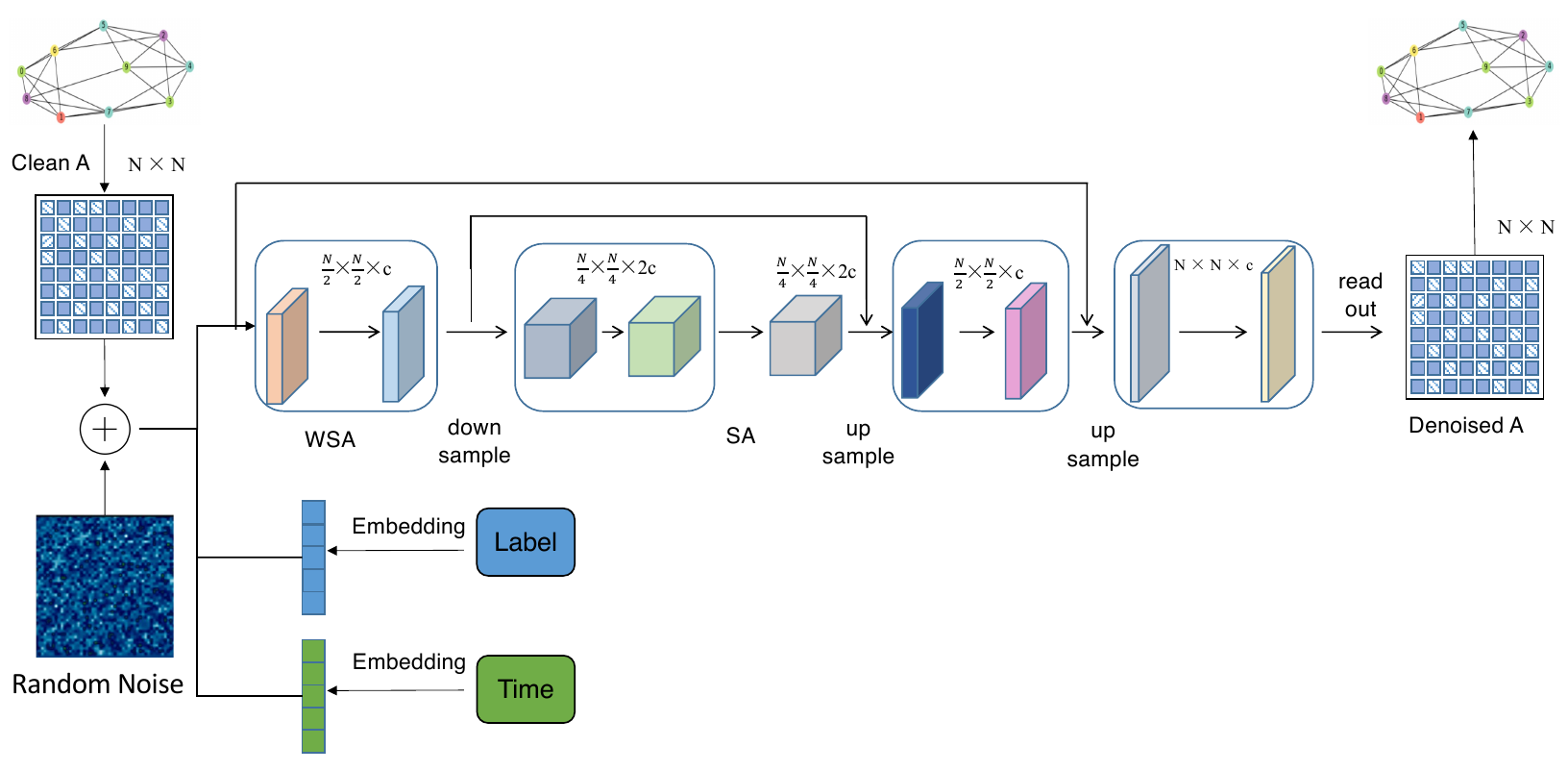}
    \caption{The overall architecture of our model, adapted from SwinGNN~\cite{yan2023swingnn}, processes graph inputs by first transforming them into noisy adjacency matrices. These matrices are then concatenated with label embeddings and passed through multiple layers of the model. After processing through these blocks, the model outputs a denoised, clean adjacency matrix. This clean matrix is subsequently transformed back into graph form, completing the cycle from input to output.}
    \label{fig:test-tp-fig}
\end{figure*}

Graph diffusion models extend diffusion models to the graph domain. An example is DiGress \cite{vignac2023digress}, which operates on discrete node types (like atom types C, N, O) and edge types (such as single, double, or triple bonds). In DiGress, introducing noise into a graph involves multiplying by a transition matrix that facilitates transitions between types. Another approach is GeoDiff \cite{geodiff}, which applies diffusion to molecular graph data represented in 3D coordinates, gradually transforming them into Gaussian noise during the forward process. The reverse process denoises a random sample into a valid set of atomic coordinates. Our target task is graph classification, so deploy a graph diffusion model that naturally facilitates the likelihood computation for a test graph, similar to the generative classifiers described for the image domain by \cite{song2021scorebased, zimmermann2021score}. The analogy to image generation motivates using SwinGNN \cite{yan2023swingnn}, which treats graphs akin to images. Given the adjacency matrix $A$ of a graph $G$, SwinGNN transforms $A$ into another adjacency matrix. The transformed adjacency matrix then undergoes a forward and backward stochastic process. Treating a graph as an image naturally allows for exact likelihood computations. The choices for the noise function in this diffusion model are $\sigma(t)=0$ and $\sigma(t) = t$, where the corresponding drift term is $f(A,t) = 0$ and the diffusion coefficient is $g(t)=\sqrt{2\dot{\sigma}(t)\sigma(t)}$. In the graph domain, Equation \ref{eq: x_forward} and Equation \ref{eq: x_backward} of the Stochastic Differential Equation (SDE) is given by:
\begin{align}
    dA_t^{+} &= \sqrt{2t}dW, \\
   dA_t^{-} &= -2t \nabla_{A} \log p_t(A)dt + \sqrt{2t} dW
\end{align}
Similar to SwinGNN and EDM \cite{yan2023swingnn, edm}, we adopt network preconditioning to improve training dynamics, and repamatrize the denoising network $D_\theta$ as $F_\theta$.

The resulting training loss is
\[
\mathbb{E}_{\sigma, A, \tilde{A}} \left[ \lambda(\sigma) \left\Vert c_s(\sigma) \tilde{A} + c_o(\sigma) F_\theta \left(c_i(\sigma) \tilde{A}, c_n(\sigma) \right) - A \right\Vert^2_{F} \right],
\]
where $\tilde{A}$ is the noisy adjacent matrix, $\sigma$ is the noisy level.
For simplicity of notation and implementation, we use a wrapper $D_\theta$ for the preconditions and reparameterization. \begin{equation}
\label{eq:D_wrapper}
    D_{\theta}(\Tilde{A}, \sigma) = c_s(\sigma) \tilde{A} + c_o(\sigma) F_\theta \left(c_i(\sigma) \tilde{A}, c_n(\sigma) \right)
\end{equation}
Then the {\em training loss objective} becomes similar to a common VDM objective 
(for more details please see the Appendix \ref{App: diffusion}): 

\begin{equation}
\label{eq:denoise_loss}
    \mathbb{E}_{\sigma, A, \tilde{A}} \left[ \lambda(\sigma) \left\Vert D_{\theta}(\Tilde{A}, \sigma) - A\right\Vert_{F}^{2} \right].
\end{equation}

\section{Training Graph Diffusion Models for Graph Classification} \label{sec:training}

\begin{figure*}
    \centering
    \includegraphics[width=\textwidth, height=0.3\textheight, keepaspectratio]{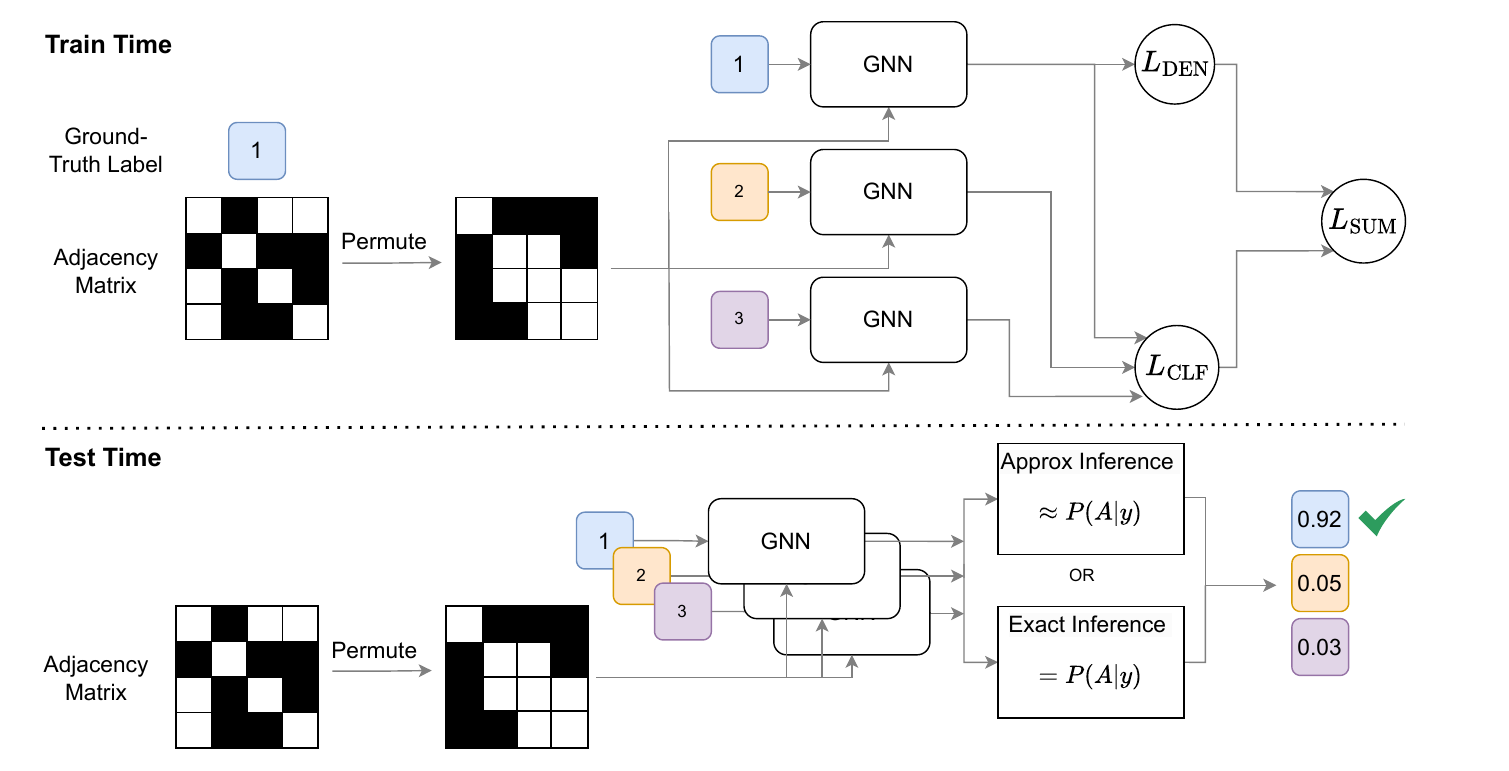}
    \caption{Overview of training and testing procedure in our method. \textit{Train Time:} The input adjacency matrix is first randomly permuted, then input to a GNN along with all possible class labels (separately, but in the same model). The loss $\denLoss$ is then computed based on the output of the ground-truth class, the $\clfLoss$ is computed based on the output of all the labels, and $\sumLoss$ is the sum of two losses. \textit{Test Time:} The adjacency matrix is permuted and input to the GNN along with all possible labels, similar to train time. Then, we compute the likelihood of the graph given each label via exact or approximate likelihood computation. We obtain the posterior probability for each label and pick the label with the highest probability. At inference time, the procedure is repeated for $P$ permutations, and the final label is decided based on majority voting (not shown for brevity).}
    \label{fig:main-fig}
\end{figure*}

Let $\mathcal{M} := \left\{(G^{(i)}, y^{(i)}) \,|\, 1 \leq i \leq m \right\}$ be a training dataset of size $m$, where the $i$-th example consists of an input graph $G^{(i)}$ and a discrete label $y^{(i)} \in \{1, 2, \dots, C\}$.
Our main goal is to build a generative classifier to classify graphs. To achieve this, we introduce a novel training objective and model checkpoint selection method, which we explain in this section.

\textbf{Training Objective}.
Consider a graph-label pair $(G, y) \in \mathcal{M}$.
Let $A$ be the adjacency matrix of the observed graph $G$. Using Bayes' theorem, we can derive a graph class probability from a class-conditional graph probability:
\begin{align} 
\label{eq:bayes_theorem}
p(y|A) = & \frac{p(A|y) \, p(y)}{\sum_{i} p(A|y_i) \, p(y_i)} = [\sft(\mathbf{L})]_y \\
\mathbf{L}_{j} := & \ln{p(A|y_j)}+ \ln{p(y_j)} \text{ for all $i \in \{1, 2, \dots, C\}$} \label{eq:softmax}
\end{align}

~\Cref{eq:softmax} follows from~\Cref{eq:bayes_theorem}, with $\sft$ denoting the softmax function. Uniform class priors $p(y_j)$ can be omitted from~\Cref{eq:bayes_theorem,eq:softmax}.

We investigate three plausible training objectives for a generative classifier model. 

\begin{align} 
    \denLoss(A,y,\theta) &:= \mathbb{E}_{\sigma, \tilde{A}} \left[\lambda(\sigma) \left \Vert D_{\theta}(\tilde{A}, y, \sigma) - A\right\Vert_{F}^{2} \right] \nonumber \\
    &\leq \ln{p_{\theta}(A|y)} \label{eq:lden} \\
    \clfLoss(A,y,\theta) &:= -\ln \bigg( \big[\sft \big(
        \denLoss(A,1,\theta), \nonumber \\
    &\quad \denLoss(A,2,\theta), \ldots, \denLoss(A,C,\theta)
        \big)\big]_y \bigg) 
        \label{eq:clfloss} \\
    \sumLoss(A,y,\theta) &:= \denLoss(A,y,\theta) +  \clfLoss(A,y,\theta). \label{eq:sumloss}
\end{align}

    \Cref{eq:lden} is based on~\Cref{eq:denoise_loss}, a variational approximation to the class-conditional log-likelihood of $A$.~\Cref{eq:clfloss} uses the approximate class-conditional graph log-probability to derive an approximate graph class log-probability via~\Cref{eq:softmax}.~\Cref{eq:sumloss} combines the generative and discriminative losses so that the model generates realistic graphs while also supporting classification. 

Note that {\em the classification objective $\clfLoss$ is a lower bound on the training set cross-entropy}. To show this, let $A^{(j)}$ be the adjacency matrix of the $j$-th graph $G^{(j)}$. Then 
\begin{align} \label{eq:cross-entropy}
& \frac{1}{m}\sum_{j=1}^{m} \ln p_{\theta}(y=y^{(j)}|A^{(j)}) \nonumber \\
 \geq   & \frac{1}{m} \sum_{j=1}^{m} \underbrace{\ln\left(\sft (\denLoss(A,1,\theta),\ldots,\denLoss(A,C,\theta))_{y^{(j)}} \right)}_{\clfLoss(A^{(j)}, y^{(j)}, \theta)}
\end{align}
\textbf{Training-time Data Augmentation with Random Permutations}.
During training, we observed that our model tended to overfit rapidly, impairing its ability to generalize to the validation or test sets. 
Since the SwinGNN architecture is not a permutation equivariant architecture (\ie, different permutations of the adjacency matrix give different outputs), one natural approach to prevent overfitting is to augment the dataset by considering random permutations of graphs.
Therefore, we randomly permute the input adjacency matrices for each training batch so the batch is trained on a set of permuted adjacency matrices.

\section{Classification Model} \label{sec:classify}

Let $G$ be a graph to be classified with adjacency matrix $A$. We use the Bayes' theorem formula in~\Cref{eq:bayes_theorem} to derive class probabilities  $p(y|A)$ from class-conditional probabilities $p(A|y)$. We investigate two basic methods for estimating $p(A|y)$. 

\textbf{Approximate Inference}. Approximate Inference uses the variational approximation to the class-conditional graph log-likelihood, as in the $\clfLoss$ of~\Cref{eq:clfloss}:
\begin{equation} \label{eq:clfprob}
    \ln{p(A|y)} \approx -\clfLoss(A,y,\theta) 
\end{equation}
\textbf{Exact Inference} One of the strengths of a score-based diffusion model is that exact likelihood computation is possible. 
In the image domain, \citet{zimmermann2021score} show how a trained class conditional score-based diffusion model with SDE can be used as zero-shot classifiers with impressive classification accuracy. 
Adapted from Equation \ref{eq:conditioned_loglikelihood} and Equation \ref{eq:conditioned_f}, given a trained class-conditional SwinGNN $D_{\theta}$, the exact likelihood computation based on ODE is as follows:
\begin{equation}
\label{eq:conditioned_loglikelihood_A}
\begin{aligned}
   \ln p(A|y) &= \ln p_T \left( A + \int_0^T \tilde{f}_\theta(A(t), t, y) \, dt \right) + \nonumber \\
   &\int_0^T \nabla \cdot \tilde{f}_\theta(A(t), t, y) \, dt,
\end{aligned}
\end{equation}
with 
\begin{equation}
\label{eq:conditioned_f_A}
   \tilde{f}_\theta(A(t), t, y) = \frac{A(t) - D_{\theta}(A(t), t, y)}{t},
\end{equation}
where $A$ is the adjacency matrix of the input graph, $A(t)$ is the adjacency matrix at time $t$ of the stochastic process, $y$ is the graph class label and $D_{\theta}(\Tilde{A}, y, t)$ is the output of the class-conditioned denoising network.


\textbf{Test-time Data Augmentation with Random Permutations} Recall that the SwinGNN model is not permutation-invariant. We  can view the class probability of a {\em graph} as the expectation of the class probability of the adjacency matrices that represent it: 
$p(G|y) \approx \mathbb{E}_{A \in \Pi(G)}[p(A|y)]$ where $\Pi(G)$ is the isomorphism class of the graph's adjacency matrices. 
Our use of permutations during training can be viewed as approximating the training graph probability by sampling from the isomorphism sets of the training graphs. 

At test time, we propose to utilize a similar permutation trick, and predict the class label based on several permutations of the graph. 
The predictions from different permutations are combined via majority vote to output the final solution. The majority vote avoids sensitivity to outlier permutations that might produce extreme probabilities. The exact working mechanism of permutation sampling is shown in Algorithm \ref{alg:Classification Sampling}. We apply permutation sampling with both approximate and exact inference.

\begin{algorithm}[tb]
   \caption{Inference: Classification with Sampling}
   \label{alg:Classification Sampling}
\begin{algorithmic}[1]
    \STATE {\bfseries Input:} Adjacency matrix $A$, number of permutations $P$
    \STATE {\bfseries Output:} Predicted label $\hat{y}$
    \STATE Initialize $\text{ClassifiedList} \leftarrow []$
    \FOR{$i=1$ {\bfseries to} $P$}
        \STATE $A_\text{i} \leftarrow \text{randomPermute}(A)$ \COMMENT{Randomly permutes the graph}
        \STATE $\text{ClassifiedList} \leftarrow \text{ClassifiedList} + [\text{Classify}(A_i))]$ \COMMENT{Classifies the permuted graph using Eq. \eqref{eq:clfprob} or \eqref{eq:conditioned_loglikelihood}}
    \ENDFOR
    \STATE \textbf{return} $\text{majorityVote}(\text{ClassifiedList})$ \COMMENT{Returns the most frequent classification}
\end{algorithmic}
\end{algorithm}

\begin{table*}[t]
\caption{Comparison of our generative method with various discriminative GNN baselines on the IMDB-BINARY and PROTEINS dataset. Number for other methods are recorded from \citet{Errica2019AFC}.}
\label{tab:compare}
\centering
\resizebox{0.8\textwidth}{!}{
\begin{tabular}{lccc}
\toprule
Method & IMDB-BINARY & IMDB-BINARY w. features & PROTEINS\\
\midrule
DGCNN~\cite{dgcnn} & 53.3 $\pm$ 5.0 & 69.2 $\pm$ 3.0  & 72.9 $\pm$ 3.5\\
DiffPool~\cite{diffpool} & 68.3 $\pm$ 6.1 & 68.4 $\pm$ 3.3 & 73.7 $\pm$ 3.5\\
ECC~\cite{ecc} & 67.8 $\pm$ 4.8 & 67.7 $\pm$ 2.8 & 72.3 $\pm$ 3.4\\
GIN~\cite{xu2018how} & 66.8 $\pm$ 3.9 & 71.2 $\pm$ 3.9 & 73.3 $\pm$ 4.0\\
GraphSAGE~\cite{graphSAGE} & 69.9 $\pm$ 4.6  & 68.8 $\pm$ 4.5 & 73.0 $\pm$ 4.5\\
Ours & \textbf{70.5} $\pm$ \textbf{5.7} &   \textbf{72.6} $\pm$  \textbf{2.1} & \textbf{75.4} $\pm$ \textbf{3.4}\\
\bottomrule
\end{tabular}
}
\end{table*}

\begin{table*}[t]
\caption{Ablation accuracies for the total of 6 design space of our training and inference combination on K-Regular, IMDB-BINARY, IMDB-BINARY with features and PROTEINS datasets. EXACT and APPROX refer to exact inference and approximate inference, respectively. Generally, larger number of permutations lead to higher accuracy, and $\clfLoss$ outperforms other strategies.}
\label{tab:ablation}
\begin{center}
\begin{small}
\begin{sc}
\resizebox{0.8\textwidth}{!}{
\begin{tabular}{lcccc}
\toprule
\shortstack{ Dataset \\ \space} & \shortstack{Inference \\(Permutations)} & \shortstack{Train w. \\ $\denLoss$}  & \shortstack{Train w. \\ $\clfLoss$}  & \shortstack{ Train w. \\ $\sumLoss$}  \\
\midrule
\multirow{3}{*}{K-regular}
& Exact (P=1)           & 66.33  & 100.00  & 100.00  \\
& Exact (P=3)           & 76.33  & 100.00  & 100.00  \\
& Exact (P=5)           & 79.00  & 100.00  & 100.00  \\
& Approx (P=5)           & 89.67  & 100.00  & 100.00   \\
& Approx (p=300)        & 100.00  & 100.00  & 100.00  \\
\midrule
\multirow{3}{*}{IMDB-BINARY} 
& Exact (P=1)         & 55.0 $\pm$ 4.29 & 67.3 $\pm$ 4.16 & 63.3 $\pm$ 4.59 \\
& Exact (P=3)         & 56.3 $\pm$ 4.36 & 67.1 $\pm$ 4.84 & 63.5 $\pm$ 5.78 \\
& Exact (P=5)         & 64.2 $\pm$ 3.31 & 67.4 $\pm$ 5.37 & 66.3 $\pm$ 4.36 \\
& Approx (p=5)        & 57.1 $\pm$ 3.81 & 69.8 $\pm$ 5.10 & 65.5 $\pm$ 4.00 \\
& Approx (p=300)      & 66.3 $\pm$ 3.61 & \textbf{70.5} $\pm$ \textbf{5.71} & 68.1 $\pm$ 5.18 \\
\midrule
\multirow{3}{*}{\shortstack{IMDB-BINARY \\ w. features}}
& Exact (P=1)         & 59.4 $\pm$ 5.26 & 68.8 $\pm$ 3.43 & 66.2 $\pm$ 3.43 \\
& Exact (P=3)         & 63.4 $\pm$ 4.36 & 71.9 $\pm$ 2.70 & 67.7 $\pm$ 2.80 \\
& Exact (P=5)         & 65.9 $\pm$ 3.27 & 71.1 $\pm$ 3.18 & 70.0 $\pm$ 4.36 \\
& Approx (p=5)        & 65.4 $\pm$ 3.90 & 69.6 $\pm$ 4.72 & 70.1 $\pm$ 3.83 \\
& Approx (p=300)      & 72.1 $\pm$ 2.21 & \textbf{72.6} $\pm$ \textbf{2.11} & 70.2 $\pm$ 4.26 \\
\midrule
\multirow{3}{*}{PROTEINS}
& Exact (P=1)         & 55.7 $\pm$ 5.85 & 62.3 $\pm$ 10.13 & 70.4 $\pm$ 3.50 \\
& Exact (P=3)         & 59.0 $\pm$ 6.88 & 62.8 $\pm$ 8.21  & 71.4 $\pm$ 3.84 \\
& Exact (P=5)         & 57.1 $\pm$ 6.27 & 62.7 $\pm$ 7.37  & 70.2 $\pm$ 4.54 \\
& Approx (p=5)        & 64.1 $\pm$ 3.29 & 74.1 $\pm$ 3.00  & 72.5 $\pm$ 4.08 \\
& Approx (p=300)              & 72.4 $\pm$ 4.44 & \textbf{75.4} $\pm$ \textbf{3.39}  & 75.4 $\pm$ 4.00 \\
\bottomrule
\end{tabular}
}
\end{sc}
\end{small}
\end{center}
\end{table*}

\section{Experiments} \label{sec: experiment}

\textbf{Setup.} We use SwinGNN as our backbone model, the choices of the diffusion schedule and parameter of the models is aligned with previous work ~\cite{edm, yan2023swingnn}, see Section \ref{sec: swinGNN} and Appendix \ref{App: diffusion}. All results are done using the 10-fold split same as \cite{Errica2019AFC}. The values are reported as the mean of the 10-fold values, and the standard deviation are reported as the standard deviation of those values. More detail in Appendix \ref{App: experiment}.

\textbf{Model Checkpoint Selection}.\label{sec:model_selection}
Our training approach includes a model checkpointing strategy: For each fold, we train the model for a predetermined number of epochs $T$, saving the model after every 100 epochs. We then choose the best model based on its accuracy on the validation set.
However, evaluating the validation set accuracy through exact inference (described in \Cref{sec:classify}) involves solving ODEs and is consumes a significant amount of computation time. For example, assessing the accuracy of a single checkpoint on the IMDB-B dataset’s validation set takes about 20 minutes per fold. While it would be beneficial to evaluate numerous checkpoints, the substantial time requirement restricts us to a limited selection.
Conversely, evaluating the variational approximation loss $\clfLoss$ from \Cref{eq:clfloss} is considerably quicker, as approximate inference is much less time-intensive compared to solving ODEs. For instance, calculating the classification loss on the validation set takes approximately 0.5 seconds. This efficiency results in a speedup of over 2000 times in checkpoint selection compared to ODE solving.
Therefore, using approximate inference allows for the evaluation of many more checkpoints, facilitating the selection of the most effective model based on performance on the validation set.

\textbf{Datasets.} We consider three datasets (1) a synthetic $K$-regular graph dataset (\ie, in each graph all nodes have the same degree $K$). Graphs fall into a category of $4$-regular and $6$-regular, and the task is a binary graph classification task. (2) 
IMDB-B~\cite{imdb_source} dataset, which consists of ego-graphs of IMDB actors/actresses. In each graph, two nodes are connected if their corresponding actors/actresses have occurred in a movie. The task is binary classification of graphs into two Genres of Action and Romance. IMDB-BINARY (FEATURES) is the IMDB-BINARY dataset that each node has the number of degrees as their features. (3) PROTEINS \cite{protein} dataset, a binary classification dataset consist of proteins that are classified as enzymes or non-enzymes. For IMDB-B and PROTEINS, we use the same split as \citet{Errica2019AFC} to ensure a fair comparison with the GNN baselines.

\textbf{Baselines.} For IMDB-BINARY, we include the same baselines as \citet{Errica2019AFC}. 


Our three training objectives (~\Cref{sec:training}) and two inference methods (~\Cref{sec:classify}) define a space of 6 possible designs. One of these designs is a zero-shot classifier baseline that has been previously applied to images. The combination of using $\denLoss$ as a training objective with exact inference for classification is analogous to the approach of~\cite{zimmermann2021score} for image classification.


\subsection{Results} \label{sec:results}

\textbf{Generative classifier is competitive with discriminative baselines.} Table \ref{tab:compare} shows the comparison of our method with various baselines. As shown in the Table, For IMDB-BINARY without node features, our method achieves better average compared to the best discriminative model (GraphSAGE), while showing higher variance. When we train with node feature, our method shows both better average and lower variance then previous SOTA (GIN). This competitive performance shows promise in generative models for classification tasks.

\textbf{More inference-time permutations improve test accuracy.} Table \ref{tab:ablation} demonstrates how increasing the number of permutations increase accuracy on all three datasets experiment, where we achieve from $2\%$ to more than $10\%$  gain across the board by increasing the number of permutations from $1$ to $5$. Furthermore, Figure \ref{fig:test-tp-fig} shows the clear trend in accuracy gain with higher number of permutations, although the gains get saturated at a certain threshold (100 in our case). Both results show the  benefit of inference-time permutation for classification.



\textbf{Approximate Inference pair with $\clfLoss$ is the best classifier combination.} Table \ref{tab:ablation} also shows that among three training objective strategies ($\clfLoss, \denLoss, \sumLoss$), using the $\clfLoss$ pair with Approximate Inference outperforms all other methods. It achieve better results for all four datasets and all other baslines.\\



\begin{figure}[htbp] 
  \centering
  \includegraphics[width=\linewidth]{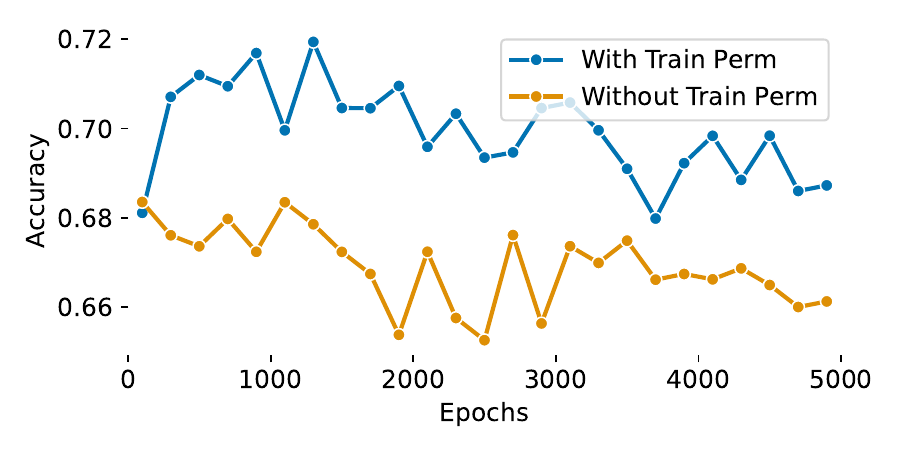}
  \caption{The mean accuracy curve during training using $\clfLoss$ on IMDB-BINARY dataset when training with or without random permutation of the adjacency matrix. Training with permutations offers significant gains in accuracy.}
  \label{fig:train-tp-fig}
\end{figure}

\begin{figure}[htbp] 
  \centering
  \includegraphics[width=\linewidth]{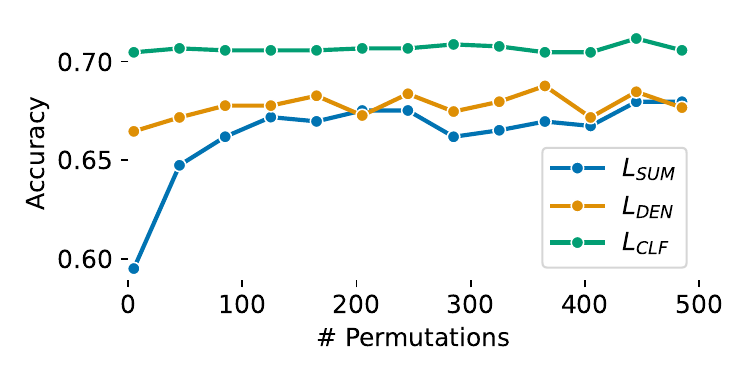}
  \caption{The accuracy curve of varying the number of permutations across three training objectives ($\clfLoss$, $\denLoss$, and $\sumLoss$) with approximate inference. In general, increasing the number of inference-time permutations improves performance, but the gains saturate around 100 permutations.}
  \label{fig:test-tp-fig}
\end{figure}



\subsection{Ablation Studies}
\textbf{Training Permutation.}
We also perform an ablation study to evaluate the impact of Training Permutation on model performance. We assess the validation set accuracy trends during training for models trained with and without Training Permutation on the IMDB-BINARY dataset using the $\clfLoss$ loss objective, as depicted in Figure \ref{fig:train-tp-fig}. The results indicate that models trained with permutation achieve higher accuracy more rapidly and maintain superior performance compared to their non-permuted counterparts, corroborating our findings in Table \ref{tab:ablation}. Nonetheless, the benefits of Training Permutation are not universally consistent; further investigations into its effects across different settings are detailed in Appendix \ref{App: training_p}.

\textbf{Inference Permutation.}
While Table \ref{tab:ablation} indicates that test accuracy generally improves with an increased number of permutations, the enhancement is not consistently linear as the number of permutations grows. Consequently, we investigate how varying the number of permutations impacts performance on the IMDB-BINARY dataset. The results for Approximate Inference are illustrated in Figure \ref{fig:test-tp-fig}. The plots reveal that a higher number of permutations does not detrimentally affect performance; however, excessively large permutation counts may be redundant, as performance gains become marginal. It is important to note that the specific effects are highly dependent on the data split and the dataset used. For results pertaining to Exact Inference, refer to Appendix \ref{App: testing_p}.

\section{Conclusion}

Our study demonstrates the significant potential of generative models for graph classification tasks, thereby expanding the applications of generative approaches beyond the text and image domains explored so far. We presented a novel training objective that preserves generative capabilities while enhancing classification performance. Our proposed inference technique, involving graph random-permutation majority voting, was shown to improve classification accuracy. Furthermore, we addressed the challenge of utilizing exact inference for score-based diffusion models during training by utilizing a variational likelihood approximation, which allows more efficient and more powerful model selection. We found that permutation-based approximate inference offers an efficient training and testing method with state-of-the-art graph classification accuracy.

A valuable direction for future work would be to consider fine-tuning a generatively trained diffusion model with our discriminative training objective. Fine-tuning combines (some of) the efficiency advantages of zero-shot generative classifiers with the accuracy of discriminatively trained generative classifiers.

Given how well established generative classifiers are for non-graph i.i.d. data, we believe that generativegraph classification is a promising new approach that will inspire new methodologies that take advantage of the unique strengths of graph generative models.



\bibliography{main}
\bibliographystyle{icml2024}

\newpage
\appendix
\onecolumn

\section{Appendix}
\subsection{Experiment}

\subsubsection{Diffusion set up} \label{App: diffusion}

For the backbone model of our experiments, we use the exact same architecture as SwinGNN \cite{yan2023swingnn}, except for the class conditioning, we add an embedding layer for each label. The class label embedding would sum up with the noise embedding before passing to each layer. And then, it would go through a reparamerterzation discussed in \cite{edm}, specifically, we have the objective function

\[
E_{\sigma, A, \hat{A}} \left[ \lambda(\sigma) \left\| c_s(\sigma) \hat{A} + c_o(\sigma) F_\theta \left(c_i(\sigma) \hat{A}, c_n(\sigma) \right) - A \right\|^2_{F} \right].
\]

where 

\[
\begin{aligned}
c_s(\sigma) &= \frac{\sigma_d^2}{\sigma_d^2 + \sigma^2}, & c_o(\sigma) &= \frac{\sigma \sigma_d}{\sqrt{\sigma_d^2 + \sigma^2}}, \\
c_i(\sigma) &= \frac{\sigma^2}{\sqrt{\sigma_d^2 + \sigma^2}}, & c_n(\sigma) &= \frac{1}{4} \ln(\sigma), \\
\lambda(\sigma) &= \frac{1}{c_o(\sigma)^2}, & \sigma_d &= 0.5, \\
\ln(\sigma) &\sim N\left(p_{\text{mean}}, p_{\text{std}}^2\right), & p_{\text{mean}} &= -1.2, \\
\sigma_{\text{min}} &= 0.002, & \sigma_{\text{max}} &= 80, & \rho &= 7, \\
S_{\text{min}} &= 0.05, & S_{\text{max}} &= 50, & S_{\text{noise}} &= 1.003, \\
S_{\text{churn}} &= 40, & N &= 256, \\
t_i &= (\sigma_{\text{max}}^{\frac{1}{\rho}} - \sigma_{\text{min}}^{\frac{1}{\rho}}) \frac{i}{N-1} + \sigma_{\text{min}}^{\frac{1}{\rho}}, \\
\gamma_i &= \begin{cases}
1 & \text{if } S_{\text{min}} \leq t_i \leq S_{\text{max}}, \\
\min\left(\frac{S_{\text{churn}}}{N}, \sqrt{2} - 1\right) & \text{otherwise},
\end{cases}
\end{aligned}
\]

\subsubsection{Datasets and Experment detail} \label{App: experiment}
For the K-regular dataset, we generate 500 4-regular graphs and 6-regular graphs. We use a split ratio of 1/1/8 for the test/validation/train sets for this synthetic dataset, and a training of 3 hours lead to us a perfect classification accuracy.
 
For IMDB-BINARY and PROTEINS, we use the same split as \citet{Errica2019AFC}, where we also use the 10-fold CV for model selection and evaluation. IMDB-BINARY with feature is IMDB-BINARY dataset with degree attach to each node. 

Since SwinGNN treat the graph as an image using its Adjacency matrix and Node attribute matrix, the model becomes more resource-intensive when the graphs grow in size, which causes efficiency issues to the model hyperparameter selection. For example, the largest graph in the Protein dataset consists of 620 nodes, but the average number of nodes is $39.1$. If we set our input dimension as 620, we need to pad \emph{all} input graphs to 620 nodes, resulting in an Adjacency matrix of size $620*620$. The SwinGNN model then needs to process \emph{all} input graphs in the size of a $620 * 620$ resolution image, which is highly inefficient, both in terms of training and inference. To overcome this issue, we select the appropriate size to be able to fit most graphs in the dataset. We omit the graphs with more nodes than our predefined cutoff. To be fair with the other methods, in inference time, for each graph in the testing and validation set that we omit, we consider the instance as an incorrect classification. This would give us a lower bound of our models test accuracy, ignoring its potential classification ability of classify large graph, in trading of efficiency. Specifically, for IMDB-BINARY, we use 128 nodes as the maximum number of nodes; For PROTEINS, we use 192. 

As training generative model is more expensive and especially in our case, which we randomly permute the training set resulting the cost of training the model more expansive, we slightly modify some of the setting. This would only make our performance sub-optimal in trade of time and resource. Specifically, We do not perform any hyper-parameter tuning for each fold, instead, we only experiment a few hyper-parameter setting on one fold, and then use the better set of hyper-parameter for all the folds. In \citet{Errica2019AFC} work, in each fold, the authors randomly do a 90/10 split three times and train three times and perform early stopping to select their model to evaluate on a test set for each fold. While in our case, it is expansive to train for three times, so instead, we only do this once. 


At inference time, to compute accuracy, we do not let the ODE solver solve the whole process from clean data to data with the maximum noisy level $\sigma(t)$ of $t=80$. Instead, we test $t=4$ is good enough, there is no need for requesting solver to solve for highly noisy data as they are less accurate and more time consuming. All experiments of this work use this setting for likelihood computation, 


\subsection{Permutation Experiments}

\subsubsection{Training Permutation} \label{App: training_p}
Note that since it is unfair and would give bad performance for models train without permutation to inference with permutation, we modify the behaviour in inference time for training without permutation. Instead of permuting $p$ times, they just classify $p$ graphs that is randomly add noise according the the training noise distribution. This process aligned with the work in \cite{secret_zero_shot}

\begin{figure}
    \centering
    \includegraphics[width=0.7\textwidth]{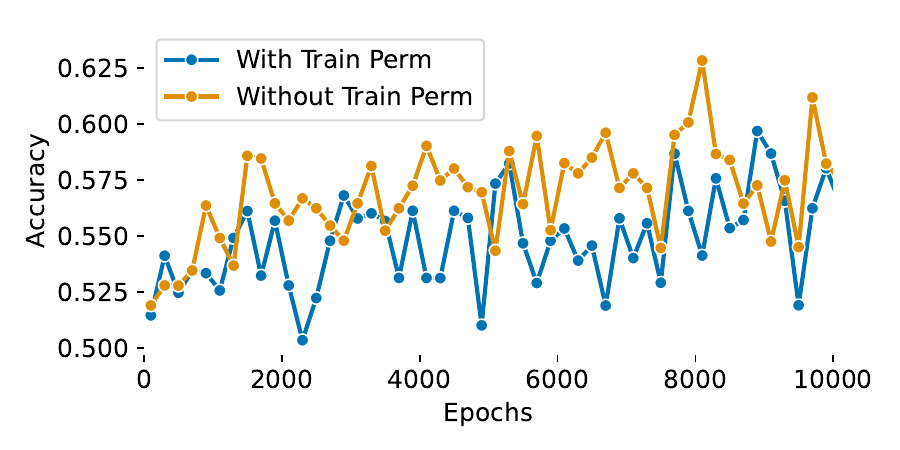}
    \caption{The mean validation accuracy curve during training using $\denLoss$ on IMDB-BINARY dataset when taining with or without random permutation of the adjacency matrix. The blue line is training without permutation, the red line is training with permutation}
    \label{fig:train-tp-a1b0-fig}
\end{figure}

\begin{figure}
    \centering
    \includegraphics[width=0.7\textwidth]{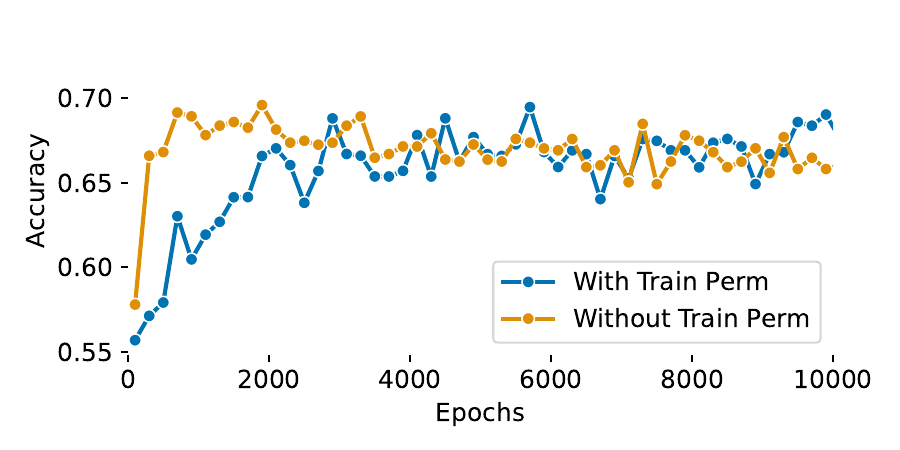}
    \caption{Similar as Figure \ref{fig:train-tp-a1b0-fig} except train on $\sumLoss$. The blue line is training without permutation, the brown line is triaing with permutaiton}
    \label{fig:train-tp-a1b1-fig}
\end{figure}

We also carried out a training permutation experiment with $\denLoss$ and $\sumLoss$ on the IMDB-BINARY dataset, as shown in Figure \ref{fig:train-tp-a1b0-fig} and Figure \ref{fig:train-tp-a1b1-fig}.

Unlike the curve in Figure \ref{fig:train-tp-fig}, which training with permutation consistently outperforms training without permutation. Figure \ref{fig:train-tp-a1b1-fig} shows comparable performance, but training without permutation achieves better accuracy faster. Figure \ref{fig:train-tp-a1b0-fig}, on the other hand, shows that training without permutation is better. Overall, $\clfLoss$ with permutation still performs the best. Their performance is poor and may result from the model having trouble leaning the features that could help in classification. Making permutation in this case makes these learning harder. This suggests that for other loss objectives, training with permutation may not always be helpful in classification,

\subsubsection{Testing Permutation} \label{App: testing_p}

\begin{figure}
    \centering
    \includegraphics[width=0.7\textwidth]{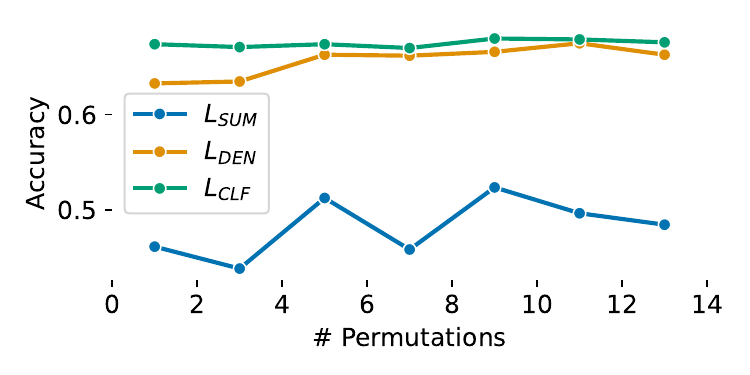}
    \caption{The test accuracy curve of the three objective using Exact inference for different number of permutation. The yellow, brown and green lines are $\clfLoss$, $\denLoss$ and $\sumLoss$ respectively}
    \label{fig: exact_test_p}
\end{figure}

We also show the inference time permutation experiment on IMDB-BINARY dataset, as shown in Figure \ref{fig: exact_test_p}. We could see that, for all three ways of training objectives $\denLoss$, $\clfLoss$, and $\sumLoss$, using the exact inference w with increasing number of permutations could benefit. Consistent with message we get from to the case in Figure \ref{fig:test-tp-fig}, it may not be beneficial to have a higher number of permutation numbers. Unlike in Approximate Inference, Exact Inference takes much more computation resource for the ODE solver (which is repeated called our neural network), it is hard to afford a permutation number as high as hundreds for Exact Inference. 

\end{document}